\documentclass{amsart} % For LaTeX2e

\usepackage{hyperref}
\usepackage{url}
\usepackage{amssymb,amsmath,amsfonts,amscd,amsthm}
\usepackage{graphicx}
\usepackage{algorithmic}
\usepackage{algorithm}
\usepackage{subfig}
\usepackage{caption}
\usepackage{url}
\usepackage{booktabs}
\newtheorem{theorem}{Theorem}[section]

\newtheorem{lemma}[theorem]{Lemma}

\newtheorem{assumption}[theorem]{Assumption}

\newcommand{\R}{\mathbb{R}}

\newcommand{\G}{\mathrm{G}}

\newcommand{\Pb}{\mathbb{P}}
\newcommand{\E}{\mathbb{E}}
\newcommand{\Sb}{\mathbb{S}}

\newcommand{\XX}{\mathcal{X}}

\newcommand{\MM}{\mathcal{M}}

\newcommand{\FF}{\mathcal{F}}

\newcommand{\mb}{\mathbf}

\newcommand{\ra}[1]{\renewcommand{\arraystretch}{#1}}

\DeclareMathOperator{\dist}{dist}

\usepackage{color}

\title{Fast Landmark Subspace Clustering}

\author{Xu~Wang and Gilad~Lerman\\
       Dept.\ of Mathematics, University of Minnesota, \\ Minneapolis, MN
       55455, USA\\ 
       \{wang1591,lerman\}@umn.edu
       }

\begin{document}

\maketitle

\begin{abstract}
Kernel methods obtain superb performance in terms of accuracy for various machine learning tasks since they can effectively extract nonlinear relations. However, their time complexity can be rather large especially for clustering tasks. In this paper we define a general class of kernels that can be easily approximated by randomization. These kernels appear in various applications, in particular, traditional spectral clustering, landmark-based spectral clustering and landmark-based subspace clustering. We show that for $n$ data points from $K$ clusters with $D$ landmarks, the randomization procedure results in an algorithm of complexity $O(KnD)$. Furthermore, we bound the error between the original clustering scheme and its randomization. To illustrate the power of this framework, we propose a new fast landmark subspace (FLS) clustering algorithm. Experiments over synthetic and real datasets demonstrate the superior performance of FLS in accelerating subspace clustering with marginal sacrifice of accuracy.
\end{abstract}

\section{Introduction}

Kernel-based learning algorithms have been highly successful for various machine learning tasks. Such algorithms typically contain two steps, first nonlinearly mapping the input data $\XX$ into a high-dimensional feature space $\FF$ and then applying linear learning algorithms on $\FF$.

When designing kernel methods, it is of primary importance to make it applicable for large datasets. In other words, kernel methods need to scale linearly w.r.t.~the number of data points. Unfortunately, most kernel methods require computation related to the kernel matrix. It scales poorly since the number of operations of merely obtaining the full kernel matrix is $O(n^2)$. Furthermore, many clustering algorithms require an eigen-decomposition of the kernel matrix. 

Motivated by this scalability problem, we define a general class of kernels and study the properties of fast randomization approximations. This is a flexible framework that allows applications in different scenarios. In the setting of subspace clustering, we propose a landmark subspace clustering algorithm and show that it is fast with marginal sacrifice in accuracy. 

\subsection{Related Works}\label{subsec:prev}

For building scalable kernel classification machines, Rahimi and Recht~\cite{DBLP:conf/nips/RahimiR07} proposed the approximation of kernels by Fourier random features and justified it by the Bochner's theorem of harmonic analysis. This technique was further analyzed and refined in other classification situations~\cite{41466, DBLP:conf/nips/RahimiR08, AISTATS2012_KarK12}. It has recently been shown to have comparable performance with deep learning~\cite{DBLP:journals/corr/LuMLGGBFCKPS14} in both scalability and accuracy. However, Hamid et al.~\cite{DBLP:conf/icml/HamidXGD14} observed that such random features for polynomial kernels contain redundent information, leading to rank-deficient matrices. Although they are able to rectify it by dimension reduction, their work indicates that applying random features as suggested directly by the Bochner's theorem may not be efficient.

In the territory of fast spectral clustering, the Nystr$\ddot{o}$m method in~\cite{DBLP:conf/cvpr/LiLKL11, DBLP:conf/icml/KumarMT09} provides a low rank approximation of the kernel matrix by sampling over its columns. The error analysis of this approximation by special sampling schemes can be found in~\cite{DBLP:conf/alt/ChoromanskaJKMM13, DBLP:journals/jmlr/DrineasM05, DBLP:journals/jmlr/KumarMT12}. Different from the Nystr$\ddot{o}$m method, the landmark-based methods~\cite{DBLP:conf/aaai/ChenC11, DBLP:conf/kdd/YanHJ09, DBLP:conf/icdm/GanSN13} first generate representative landmarks to summarize the underlying geometry of a dataset and then create new columns from these landmarks instead of sampling the original columns. Comparing with generating random Fourier features or subsampling the original columns, the landmark-based methods need less columns and thus are more efficient for spectral clustering. 

\subsection{Contributions of This Paper}

First of all, we propose a randomization procedure that applies to a relatively large class of kernels, in particular, larger than classes treated by previous works (see Section~\ref{subsec:prev}). Second of all, we develop a kernel-based algorithm for fast approximated subspace clustering for a dataset of $n$ points with $K$ clusters. Given an approximation by $D$ landmarks, the algorithm requires $O(KnD)$ running time. Third of all, in the context of kernel approximations via Fourier random features and landmarks, this is the first work that provides a bound of the $L_2$-error between the original and approximated eigenvectors used in the clustering procedure. This bound is independent of $n$. We remark that Rahimi and Recht~\cite{DBLP:conf/nips/RahimiR07} only provided a bound of the difference between the corresponding entries of the original and approximated kernels.

\subsection{Organization of This Paper}

Section~\ref{sec:kernel} defines a set of kernels that fit our approximation scheme and describes a randomized approximation procedure for such kernels. Section~\ref{sec:fsc} introduces the fast landmark subspace (FLS) clustering algorithm. Section~\ref{sec:theory} shows that good approximation errors for both the kernel matrix and the eigenvectors of the normalized kernel matrix can be obtained by changing the number of landmarks with no dependence on the number of points. This analysis applies not only to FLS but also to other clustering algorithms associated with the different kernels introduced in Section~\ref{sec:kernel}. In Section~\ref{sec:experiment}, we compare the proposed FLS with state-of-the-art fast subspace clustering algorithms over various datasets.

\section{Kernels}\label{sec:kernel}

\subsection{Construction of Kernels}

In this section we introduce a class of kernels. Let $(X, \mu_X)$ and $(Y, \mu_Y)$ be two measurable spaces and $f$ be a bounded continuous map:
$$
f : \quad X \times Y \rightarrow \R,
$$
which is $L_2$-integrable w.r.t. to the product measure $\mu_X\times \mu_Y$. This map induces an embedding 
\begin{equation}\label{eq:phi}
\phi: \, x\in X \mapsto f(x,\pmb{\cdot}) \in L^2(Y, d\mu_Y).
\end{equation}
Moreover, if we define $k(x_1, x_2)$ on $X\times X$ as follows:
\begin{equation}\label{eq:ker}
k(x_1, x_2) = \int_{y\in Y} f(x_1, y) f(x_2, y) d\mu_Y(y),
\end{equation}
then the Fubini's theorem and~\cite[Page 6]{fasshauer_positive_2011} imply that $k(x_1, x_2)$ is a Mercer's kernel. Indeed, for any $u(x) \in L^2(X, d\mu_X)$, the Mercer's condition is satisfied:
\begin{equation*}
\begin{aligned}
\int_{x_1, x_2\in X} u(x_1)& k(x_1, x_2) u(x_2) d\mu_X(x_1) d\mu_X(x_2) \\  
 & = \int_{y\in Y} \left( \int_{x\in X} f(x, y)u(x)d\mu_X(x) \right)^2 d\mu_Y(y) \geq 0.
\end{aligned}
\end{equation*}
We now show that different types of kernels can be constructed in this way.

\paragraph{\bf Example I} 

The first example shows that different kernels can be obtained by varying the measure $\mu_Y$. Let $X = \R^d$ and $\mu_X$ be the Lebesgue measure. Let $Y = \R^d \times [0, 2\pi]$ and $\mu_Y$ be the product of the Gaussian measure $N(\mb{0}, 1/\sigma^2 \mb{I})$ and the uniform measure $U([0, 2\pi])$. If we define 
\begin{equation}\label{eq:f}
f (\mb{x}, (\mb{w}, t) ) = \sqrt{2}\cos(\mb{w}^T \mb{x} + t),
\end{equation}
then the corresponding kernel $k(\mb{x}_1, \mb{x}_2)$ defined in~\eqref{eq:ker} is the usual Gaussian kernel with variance $\sigma^2$. This formulation is a special case of the following Bochner's theorem (see also~\cite{DBLP:conf/nips/RahimiR07}) from harmonic analysis.
\begin{theorem}[Bochner~\cite{MR1038803}]\label{thm:bochner}
A continuous shift-invariant kernel $k(\mb{x}, \mb{y}) = k(\mb{x}-\mb{y})$ on $\R^d$ is positive-definite if and only if $k(\delta)$ is the Fourier transform of a non-negative measure.
\end{theorem}

The implication of Theorem~\ref{thm:bochner} is that every positive-definite shift-invariant kernel $k(\mb{x}_1, \mb{x}_2)$ on $\R^d$ can be written as the expectation (integral) 
$$
\E_{\mu \times U([0, 2\pi])} (f(\mb{x}_1, (\mb{w}, t)) f(\mb{x}_2, (\mb{w}, t))),
$$
where $f$ is as in~\eqref{eq:f} for some non-negative measure $\mu$. In other words, the set of kernels defined in~\eqref{eq:ker} by choosing different measures $\mu_Y$ encompass all positive-definite shift-invariant kernels on $\R^d$.

\paragraph{\bf Example II}
The second example shows that this new formulation of kernels provides a theoretical foundation for the landmark-based methods~\cite{DBLP:conf/aaai/ChenC11, DBLP:conf/kdd/YanHJ09, DBLP:conf/icdm/GanSN13}. Indeed, let $X, Y$ be $\R^d$, $\mu_X$ be the Lebesgue measure and $\mu_Y$ be the probability measure from which the observed dataset is sampled. If we define 
\begin{equation}\label{eq:fII}
f (\mb{x}, \mb{y} ) = (2\pi\sigma^2)^{-k/2}e^{-\|\mb{x}-\mb{y}\|_2^2/(2\sigma^2)},
\end{equation}
then the corresponding kernel $k(\mb{x}_1, \mb{x}_2)$ on $X$ defined in~\eqref{eq:ker} is the kernel implicitly used in landmark-based methods. Thus the theoretical analysis in Section~\ref{sec:theory} applies to these methods. We note that if one picks $f$ as above and $\mu_Y$ be the Lebesgue measure, one obtains the Gaussian kernel on $X$. 

\subsection{Alternative Interpretation}\label{subsec:alter}

The basic interpretation of this construction of kernels is that $f(x,\pmb{\cdot})$ can be taken as an infinite dimensional representation of $x$ in $L^2(Y, d\mu_Y)$ and $k(x_1, x_2)$ be the corresponding inner product. Alternatively speaking, each point $y\in Y$ provides a similarity score $f(x_1, y) f(x_2, y)$ for all pairs $(x_1, x_2)\in X\times X$. The kernel is an average of all such scores provided by the measurable space $Y$. This alternative interpretation provides an insight on the choice of $\mu_Y$.

In Example II of Section~\ref{sec:kernel}, we mentioned that the integration of~\eqref{eq:fII} over the underlying distribution of a given dataset leads to landmark-based kernels and the integration over the Lebesgue measure leads to the usual Gaussian kernels. It is evident that landmark-based kernels are more efficient since they use the average scores of landmarks, which summarize the underlying geometry and thus provide more relevant similarity scores than points randomly chosen from the whole $\R^d$.

\subsection{Randomized Approximation}\label{subsec:randommap}

In this section we present a randomized mapping $\psi$ to approximate $\phi$. We explicitly embed each $x\in X$ into an Euclidean space $\R^D$ using a randomized mapping:
$
\psi :  x \rightarrow \psi(x) \in \R^D
$
where the inner product in $\R^D$ approximates the kernel function. That is,
$$
k(x_1, x_2) = \langle \phi (x_1), \phi (x_2) \rangle_{\mu_Y} \approx \langle \psi (x_1), \psi (x_2) \rangle.
$$
Let $\psi_{y} (x) = f(x, y)$. Then $\E_{\mu_Y}[\psi_{y}(x_1) \psi_{y}(x_2)] = k(x_1,x_2)$ for all $x_1, x_2 \in \R^d$ by definition. We randomly draw $D$ points $y_i \in Y$ according to the probability measure $\mu_Y$ and define a random map:
\begin{equation}\label{eq:psi}
\psi : \quad x \in X \rightarrow \frac{1}{\sqrt{D}}[\psi_{y_1}(x), \ldots, \psi_{y_D}(x)]^T \in \R^D.
\end{equation}

\section{Fast Subspace Clustering}\label{sec:fsc}

\subsection{Subspace Kernels}

Different kernels can be obtained by taking a suitable measurable space $Y$ in~\eqref{eq:ker}. In this section we show how this idea can be applied to the subspace clustering task, leading to a fast subspace clustering algorithm. Let the measurable space $X$ be the sphere $\Sb^d$ (after normalizing the data) and the measurable space $Y$ be the Grassmannian $\G(d, l)$ of $l$-dimensional subspaces with a measure $\mu_Y$. We define the function
$$
f : (x, L) \in X\times Y \mapsto \exp(-\dist(x, L)^2/\sigma^2) \in \R,
$$
and the subspace kernel
\begin{equation}\label{eq:kernel}
k(x_1, x_2) = \int_{L \in Y} f(x_1, L) f(x_2, L) d\mu_Y(L).
\end{equation}
We explain why this can be a good kernel for spectral subspace clustering in Section~\ref{subsec:alter} below and further illustrate its power in the numerical section. 

\subsection{Choice of $\mu_Y$}\label{subsec:subker}

Lemma~\ref{lm:rotation} shows that the uniform measure on the Grassmannian $\G(d, l)$ is not a good choice. Indeed, the resulting kernel only has distance information between points and thus is a shift-invariant kernel in Example I of Section~\ref{sec:kernel}, which is not capable to detect the linear structures.

\begin{lemma}\label{lm:rotation}
If $\mu_Y$ is the uniform measure on $\G(d, l)$, then $k(x_1, x_2)$ is a function $g(\dist(x_1,x_2))$ depending only on the Euclidean distance between $x_1$ and $x_2$. 
\end{lemma}

Section~\ref{subsec:alter} provides an answer for how to pick $\mu_Y$ for subspace kernels~\eqref{eq:kernel} given a subspace clustering task. Indeed, we note that the true underlying subspaces provide more relevant scores for the clustering task. A true subspace assigns $1$ to a pair $(x_1, x_2)$ if both of them belong to this subspace and a small score if otherwise. Therefore, a good choice of $\mu_Y$ is a measure supported on a neighborhood of the underlying subspaces of a given dataset in $\G(d, l)$.

\paragraph{Landmark Subspace} Empirically the true subspaces are unknown and $\mu_Y$ can not be explicitly specified. Therefore the best we can do is to generate a set of subspaces $\{L_i\}_{i=1}^D$ that are close to the true subspaces in order to have a good approximation $k(x_1, x_2) \approx \frac{1}{D}\sum_{i=1}^D f(x_1, L_i) f(x_2, L_i).$
This set of subspaces is generated as follows. We first pick $D$ landmark points for the dataset by random sampling or $K$-means. For each landmark point, a local best fit flat is found by selecting an optimal neighborhood radius according to~\cite{DBLP:journals/corr/abs-1010-3460}. These local flats approximate the true subspaces and are called landmark subspaces.

\subsection{The FLS Algorithm}

With the kernel approximation in Section~\ref{subsec:randommap}, we can easily define the approximated normalized kernel matrix $\hat{\mb{L}}_{n, D}$ for a given dataset $X = \{\mb{x}_i\}_{i=1}^n$. Let
$$
\hat{\mb{W}} = \psi(X)^T \psi(X) = [\psi(\mb{x}_1) \cdots \psi(\mb{x}_n)]^T [\psi(\mb{x}_1) \cdots \psi(\mb{x}_n)]
$$ 
and the degree matrix $\hat{\mb{D}}$ be a diagonal matrix with the diagonal entries $\hat{D}_{ii} = \langle \psi(\mb{x}_i), \sum_{j=1}^n \psi(\mb{x_j}) \rangle$. Then we define
$$
\hat{\mb{L}}_{n,D} = \hat{\mb{D}}^{-1/2} \hat{\mb{W}} \hat{\mb{D}}^{-1/2}.
$$
The clustering is based on the top $K$ eigenvectors of $\hat{\mb{L}}_{n,D}$. At first glance, there seems no reason to consider $\hat{\mb{L}}_{n,D}$ since its construction requires $O(n^2D)$ operations. The essential idea of our randomized procedure is to provide a random low-rank decomposition of the normalized kernel matrix. The task of obtaining the top $K$ eigenvectors of $\hat{\mb{L}}_{n,D}$ is transformed to finding the top $K$ singular vectors of $\psi(X)\hat{\mb{D}}^{-1/2}$ since they are the same by definition. This can be done in $O(KnD)$. We formulate the algorithm as follows:

\begin{algorithm}[H]
\caption{Fast landmark subspace clustering (FLS)}
\begin{algorithmic}
\REQUIRE $X=\{\mb{x}_1, \mb{x}_2,\ldots, \mb{x}_N\}\in \R^d$: data, $d$: dimension of subspaces, $D$: number of landmark subspaces, $K$: number of clusters, $\sigma$: scaling parameter for kernel, $T, S$: default parameters for local best-fit flats.
\ENSURE Index set $\{g_i\}_{i=1}^N$ for $K$ partitions such that $x_i\in X$ belongs to group with label $g_i$.\\
\textbf{Steps}:\\
$\bullet$ Generate $D$ landmark points $\mb{y}_j$ for $X$ by either random sampling or applying $k$-means (see~\cite{DBLP:conf/aaai/ChenC11}).\\
$\bullet$ Find the local best-fit flat $L_j$ (e.g., the landmark subspace) for each landmark point (cf., Algorithm~2 in~\cite{DBLP:journals/corr/abs-1010-3460}).\\
$\bullet$ Compute $\psi : \mb{x}_i \rightarrow [f(\mb{x}_i, L_1), \ldots, f(\mb{x}_i, L_D)]^T, \quad \forall 1\leq i\leq n$. \\
$\bullet$ Compute $\hat{D}_{ii} = \langle \psi(\mb{x}_i), \sum_{j=1}^n \psi(\mb{x_j}) \rangle$. \\
$\bullet$ Find the top $K$ singular vectors $\{\hat{\mb{v}}_i\}_{i=1}^K$ for $\psi(X)\hat{\mb{D}}^{-1/2}.$ \\
$\bullet$ Normalize each row of $[\hat{\mb{v}}_2,\cdots, \hat{\mb{v}}_k]$ and call $K$-means on the rows.

\RETURN A cluster label $g_i$ for each point $\mb{x}_i$.

\end{algorithmic}
\end{algorithm}

\section{Theoretical Analysis}\label{sec:theory}

The main goal of the theoretical analysis is to bound the approximation errors of the kernel matrix entries and eigenvectors for the normalized kernel matrix. We show that the errors depend only on $D$. In other words, the number of landmarks $D$ does not have to increase as the dataset size $n$ increase in order to achieve a fixed level of approximation precision. This rigorously justifies that the cost $O(KDn)$ of our algorithm is indeed linear in $n$ without a hidden factor of $n$ in $D$.

\subsection{Uniform Convergence}\label{subsec:uniform}

Hoeffding's inequality guarantees the pointwise exponentially fast convergence in probability of $\langle \psi(x_1), \psi (x_2) \rangle = \frac{1}{D}\sum_{k=1}^D \psi_{y_k}(x_1) \psi_{y_k}(x_2)$ to $k(x_1,x_2)$. That is, 
$$
\mathrm{Pr}[ | \psi(x_1)^T \psi(x_2) - k(x_1, x_2) | \geq \epsilon ] \leq 2 \exp(-D\epsilon^2 /4).
$$ 
Recall that $f$ is the function defining the kernel $k(x_1, x_2)$. If $f$ and its derivatives in $x$ are bounded by a constant $C_f$, we have the following stronger result that asserts the uniform convergence over a compact subset of X. Theorem~\ref{thm:D} directly generalizes a similar argument in Claim~1 of~\cite{DBLP:conf/nips/RahimiR07} for more general manifolds $X$ and the new class of kernels. For the completeness, we include its proof in the appendix.

\begin{theorem}\label{thm:D}
Let $\MM$ be a compact subset of $X$. Then for the mapping $\psi$ defined in~\eqref{eq:psi},
$$
\mathrm{Pr}\left[ \sup_{\mb{x}, \mb{y} \in \MM} | \psi(\mb{x})^T \psi (\mb{y}) - k(\mb{x},\mb{y}) | \geq \epsilon \right] \leq C_{d, f}\exp(-D\epsilon^2/(16d+8))/\epsilon^2,
$$
where $C_{d, f}$ is a constant depending on $d$ and $f$ and the covering number of $\MM$. Furthermore, $\sup_{\mb{x}, \mb{y} \in \MM} | \psi(\mb{x})^T \psi (\mb{y}) - k(\mb{x},\mb{y}) | \leq \epsilon$ with any constant probability when $D = \Omega \left( \frac{1}{\epsilon^2} \log(\frac{1}{\epsilon^2})\right)$.
\end{theorem}

We denote the kernel matrix by $\mb{W}$. Given a dataset $X = \{\mb{x}_i\}_{i=1}^n$, we recall that the approximation is given by
$$
\hat{\mb{W}} = \psi(X)^T \psi(X) = [\psi(\mb{x}_1) \cdots \psi(\mb{x}_n)]^T [\psi(\mb{x}_1) \cdots \psi(\mb{x}_n)].
$$ 
Theorem~\ref{thm:D} states that the entrywise approximation quality of $\mb{W}$ by $\hat{\mb{W}}$ is independent of n. This immediately indicates a low-rank approximation of $\mb{W}$ as long as $D \ll n$. 

\subsection{Convergence of Eigenvectors}\label{subsec:eigen}

In many cases such as spectral clustering, the stability of eigenvectors is desired for a matrix approximation. For simplicity, we assume there are two clusters and consider the stability of the second largest eigenvector (the stability of the first eigenvector is trivial). The reasoning for top $K$ eigenvectors is similar. Given a dataset $X = \{\mb{x}_i\}_{i=1}^n$, we recall that the normalized kernel matrix is defined as 
$$
\mb{L}_n = \mb{D}^{-1/2} \mb{W} \mb{D}^{-1/2},
$$
where $\mb{D}$ is diagonal with $D_{ii} = \sum_{j} W_{ij}$, and its approximation defined as
$$
\hat{\mb{L}}_{n,D} = \hat{\mb{D}}^{-1/2} \hat{\mb{W}} \hat{\mb{D}}^{-1/2},
$$
where $\hat{D}_{ii} = \langle \psi(\mb{x}_i), \sum_{j=1}^n \psi(\mb{x_j}) \rangle$. In this section, we show the convergence of the second largest eigenvector of $\hat{\mb{L}}_{n, D}$ to that of $\mb{L}_n$ uniformly over $n$ as $D\rightarrow \infty$. The spectral convergence related to $\hat{\mb{W}}$ and $\mb{W}$ follows similarly. In the following analysis, we make the assumption below on the kernel $k(x_1, x_2)$:
\begin{assumption}
$k(x_1, x_2)$ is bounded from below and above by constants $l, u$ ($0 < l < u$) on the domain from which the dataset is sampled.
\end{assumption}
We remark that this is also an essential assumption in proving the consistency of spectral clustering in~\cite{MR2396807} and Gaussian kernels satisfy this assumption over any compact set in $\R^d$.

Theorem~\ref{thm:D} implies that $\hat{\mb{L}}_{n,D}$ entrywisely converges to $\mb{L}_n$ as $D \rightarrow \infty$. Therefore, the second largest eigenvector $\hat{\mb{v}}_{1,n}$ of $\hat{\mb{L}}_{n,D}$ converges to the second largest eigenvector $\mb{v}_{1,n}$ of $\mb{L}_n$ as $D \rightarrow \infty$ for each fixed $n$. In the following, we show that this convergence is uniform over $n$. That is, the approximation error depends only on $D$, even as $n \rightarrow \infty$.

The main idea is as follows. We first show that $\hat{\mb{L}}_{n,D}$ is a small pertubation of $\mb{L}_n$ in operator $L_2$-norm in Lemma~\ref{lm:l2}. Then we show that the first nonzero eigenvalue of $\mb{L}_n$ has strictly positive distance from the rest spectrum for all $n$ in Lemma~\ref{lm:firsteig}. This implies that a small perturbation of $\mb{L}_n$ does not significantly change the corresponding eigenvector $\mb{v}_{1,n}$. 

We now make the above arguments precise. To begin with, we assume the following probabilistic model. Suppose $\mb{x}_i$ in $X$ are i.i.d.~sampled from the unit ball $B(\mb{0},1) \subset \R^d$ according to some probability measure $P$. 

We denote $H_1 = (\mb{D}^{-1/2} - \hat{\mb{D}}^{-1/2}) \mb{W} \mb{D}^{-1/2}$, $H_2 = \hat{\mb{D}}^{-1/2} ( \mb{W} - \hat{\mb{W}}) \mb{D}^{-1/2}$ and $H_3 = \hat{\mb{D}}^{-1/2} \hat{\mb{W}} (\mb{D}^{-1/2} - \hat{\mb{D}}^{-1/2})$. It is easy to see that 
$$
\hat{\mb{L}}_{n,D} = \mb{L}_n - H_1 - H_2 - H_3.
$$
The first task is to show that the $L_2$-norms $\| H_i\|_2$ are small uniformly over $n$. We note that $\mathrm{diam}(B(\mb{0}, 1)) = 2$. Theorem~\ref{thm:D} implies that given any $\delta >0$ and any probability $p>0$ with $D :=  \frac{c_p}{\delta^2 l^2} \log \frac{1}{\delta l}$ where $c_p$ is a constant depending only on $p$, the entrywise error $|W_{ij} - \hat{W}_{ij}| \leq \delta l$ with probability $p$. The first task is to show that $\|H_i\|_2, i=1,2,3$ are small uniformly over $n$. This is formulated as follows and proved in the Appendix.

\begin{lemma}\label{lm:l2}
If $\delta>0, p>0$, $D :=  \frac{c_p}{\delta^2 l^2} \log \frac{1}{\delta l}$, where $c_p$ is a constant depending only on $p$, then 
$\|H_1\|_2 \leq  \frac{u}{2l} (1-\delta)^{-3/2}\delta$, $\|H_2\|_2 \leq  (1-\delta)^{-1/2} \delta$ and $\|H_3\|_2 \leq  \frac{1}{2(1-\delta)^2} (\frac{u}{l}+\delta) \delta$ with probability $p$, for all $n$.
\end{lemma}

Lemma~\ref{lm:l2} shows that $\hat{\mb{L}}_{n,D}$ is a small perturbation of $\mb{L}_n$ in $L_2$-norm uniformly over $n$. Now we analyze further the relation between their spectrums. Let $\sigma(\mb{L}_n)$ and $\lambda_{1,n}$ be the spectrum and the second largest eigenvalue of $\mb{L}_n$.  Lemma~\ref{lm:firsteig} below shows that $\lambda_{1,n}$ is sufficiently separated from the rest spectrum of $\mb{L}_n$.
\begin{lemma}\label{lm:firsteig}
Let $c >0$ be a constant depending on the generating probability $P$ of the dataset $X$ and the kernel $k(x_1, x_2)$. For any constant probability $p$, there exists $N_p$ such that 
$$
\dist(\lambda_{1,n}, \{0\}\cup \sigma(\mb{L}_n) \backslash \{\lambda_{1,n}\}) \geq c, \quad \forall n\geq N_p.
$$
\end{lemma}

Lemma~\ref{lm:firsteig} together with the Davis-Kahan theorem~\cite{MR0264450} (see also~\cite[Theorem~1]{variantDK}) indicates that $\lambda_{1, n}$ and $\mb{v}_{1,n}$ are robust (not mixed with other eigenvectors) under a small pertubation of $\mb{L}_n$. Indeed, we have
$$
\|\hat{\mb{v}}_{1,n} - \mb{v}_{1,n}\|_2 \leq \frac{\|H_1 + H_2 + H_3\|_2}{c} \leq C \delta,
$$
for some constant $C>0$ and $\delta < 1/2$. Thus Theorem~\ref{thm:uniform} below follows.

\begin{theorem}\label{thm:uniform}
For any $0<\delta < 1/2$ and $0<p\leq 1$, there are constants $c_p$ and $N_p$ such that if $D = \frac{c_p}{\delta^2 l^2} \log \frac{1}{\delta l}$ and $n>N_p$, then
$$
\|\hat{\mb{v}}_{1,n} - \mb{v}_{1,n}\|_2 \leq C \delta,
$$
with probability at least $p$. Here $C$ depends on the gap of top eigenvalues and the lower and upper bounds of the kernel $k(x_1, x_2)$ on the compact domain. 
\end{theorem}

If there are $K$ clusters, we need to consider first $K-1$ nonzero eigenvectors $\{\mb{v}_{i,n}\}_{i=1}^{K-1}$ and their perturbations $\{\hat{\mb{v}}_{i,n}\}_{i=1}^{K-1}$. One can use similar arguments as above to prove the robustness of $\{\mb{v}_{i,n}\}_{i=1}^{K-1}$. We note that the approximation error of eigenvectors is controlled by the number of landmarks $D$.

\section{Experiment}\label{sec:experiment}

We compare FLS with the following algorithms: Local Best-fit Flats (LBF)~\cite{DBLP:journals/corr/abs-1010-3460}, Sparse Subspace clustering (SSC)~\cite{DBLP:journals/pami/ElhamifarV13}, Scalable Sparse Subspace Clustering (SSSC)~\cite{DBLP:conf/cvpr/00010Y13}. LBF and SSSC are algorithms with the state-of-the-art performance in speed with reasonable high accuracy. Their comparison with other subspace clustering algorithms can be found in~\cite{DBLP:journals/corr/abs-1010-3460, DBLP:conf/cvpr/00010Y13}. Yet, there is no direct comparison between them. SSC is a popular but slow algorithm, which is included as a contrast for accuracy. We report the time cost in seconds and measure the accuracy of these algorithms by the ratio of correctly-clustered points with outliers excluded. That is,
$$
\mathrm{rate} = \frac{\# \text{ of correctly-clustered inliers}}{\#\text{ of total inliers}} \times 100\%.
$$ 
When we fail to obtain a result due to the exceeding of memory limits, we report it as N/A.

\subsection{Synthetic Data}

FLS was tested and compared with other algorithms for artificial data with various subspace dimensions and levels of outliers. We use the notation $(d_1,\ldots, d_K)\in \R^D$ to denote the model of $K$ subspaces in $\R^D$ of dimensions $d_1,\ldots, d_K$. Given each model, we repeatly generate $10$ different sample sets according to the code in~\cite{synthetic_web}. More precisely, for each subspace in the model, 250 points are first created by drawing uniformly at random from the unit disk of that subspace and then corrupted by Gaussian noise (e.g., from $N(\mb{0}, 0.05^2\mb{I}_{D\times D})$ on $\R^D$). Then the whole sample set is further corrupted by adding $5\%$ or $30\%$ uniformly distributed outliers in a cube of side-length determined by the maximal distance of the former generated data to the origin. For each model, we repeat the experiment on 10 sample sets and the average accuracy (running time) are reported in Table~\ref{table:synthetic5} and~\ref{table:synthetic30} for outlier levels $5\%$ and $30\%$ respectively.

\begin{table}[htb!]\centering
  \caption{Average accuracy (time (s)) for outlier level $5\%$.}\label{table:synthetic5} 
  \hfill \\
  \ra{1.3}
  \begin{tabular}{@{}ccccc@{}}\toprule
     & $(2,2)\in\R^{6}$ & $(4,5,6)\in\R^{10}$ & $(5,6,7)\in\R^{20}$ & $(3,4,5,6,7)\in\R^{80}$ \\
    \midrule 
    FLS & {\bf 0.99} (0.81) & {\bf 0.98} ({\bf 0.81}) & {\bf 1.00} ({\bf 1.0}) & {\bf 1.00} ({\bf 4.7}) \\
    LBF & {\bf 0.99} ({\bf 0.65}) & 0.97 (0.98) & 1.00 (1.3) & 0.98 (8.7) \\
    SSC & 0.92 (140) & 0.55 (290) & 0.98 (330) & 0.95 (160) \\
    SSSC & 0.89 (2.6) & 0.62 (4.2) & 0.83 (6.8) & 0.73 (6.4) \\
    \bottomrule
  \end{tabular}
\end{table}

\begin{table}[htb!]\centering
  \caption{Average accuracy (time (s)) for outlier level $30\%$.}\label{table:synthetic30} 
  \hfill \\
  \ra{1.3}
  \begin{tabular}{@{}ccccc@{}}\toprule
     & $(2,2)\in\R^{6}$ & $(4,5,6)\in\R^{10}$ & $(5,6,7)\in\R^{20}$ & $(3,4,5,6,7)\in\R^{80}$ \\
    \midrule 
    FLS & {\bf 0.99} (0.69) & {\bf 0.98} ({\bf 0.99}) & {\bf 1.00} ({\bf 1.2}) & {\bf 0.99} ({\bf 7.9}) \\
    LBF & {\bf 0.99} ({\bf 0.61}) & {\bf 0.98} (1.0) & 1.00 (1.6) & 0.98 (11) \\
    SSC & 0.91 (210) & 0.59 (360) & 0.84 (260) & 0.73 (760) \\
    SSSC & 0.76 (2.6) & 0.44 (4.1) & 0.47 (6.2) & 0.40 (6.9) \\
    \bottomrule
  \end{tabular}
\end{table}

%\subsection{Trade-off of speed and accuracy}
%We note that for FLS, LBF and SSSC, it is possible to have a trade-off between speed and accuracy by tuning the parameters (e.g., the number of landmarks $p$). A larger $p$ usually leads to better accuracy but slower speed. For LBF, it is a bit tricky to get the trade-off since there are two parameters affect the speed and accuracy, namely, the number of candidate subspaces and the number of iteration steps in the greedy optimization. We did a grid search to find paramters that give the optimal trade-off. SSC does not have a mechanism for the trade-off, therefore, is not plotted in Figure~\ref{fig:trade_off}.
%
%\begin{figure}[!htb]
%  \centering
%  \subfloat[Angle Filtering]
%  {\includegraphics[width=0.4\linewidth]{sample.png}
%  \label{fig:angle.threshold}}
%  \subfloat[Intersection]
%  {\includegraphics[width=0.4\linewidth]{sample.png}
%  \label{fig:intersection}}
%  \caption{Trade-off of speed and accuracy}
%  \label{fig:trade_off}
%\end{figure}

\subsection{MNIST Dataset}

We test the four algorithms on the MNIST dataset of handwritten digits. We work on the training set of $60000$ $28\times 28$ images of the digits $0$ through $9$. We pick images of a combination of several digits and apply PCA to reduce the dimension to $D=80$. We choose a fixed subspace dimension $d=3$ and the correct number of clusters. The clustering rate is reported in Table~\ref{table:mnist}. We note that $[1-9]$ stands for $[1,2,3,4,5,6,7,8,9]$ in the table.

\begin{table}[htb!]\centering
  \caption{Average accuracy (time (s)) for MNIST dataset.}\label{table:mnist} 
  \hfill \\
  \ra{1.3}
  \begin{tabular}{@{}ccccccc@{}}\toprule
      & $[1,7]$ & $[4,9]$ & $[2,4,8]$ & $[3,6,8]$ & $[2,5,7,8,9]$ & $[1-9]$\\
    \midrule 
    FLS & {\bf 0.99} (15) & {\bf 0.93} (13) & {\bf 0.97} (20) & {\bf 0.95} (19) & {\bf 0.81} (38) & {\bf 0.54} (111)\\
    LBF & 0.65 ({\bf 6}) & 0.57 ({\bf 7}) & 0.44 ({\bf 12}) & 0.57 ({\bf 12}) & 0.36 ({\bf 27}) & 0.25 ({\bf 85})\\
    SSC & 0.76 (18993) & 0.66 (14358) & 0.94 (45683) & 0.86 (47448) & N/A & N/A\\
    SSSC & 0.89 (33) & 0.65 (27) & 0.95 (34) & 0.90 (37) & 0.77 (70) & 0.50 (152)\\
    \bottomrule
  \end{tabular}
\end{table}

\subsection{Other Datasets}

We also tried the four algorithms on the Extended Yale B (ExtendYB)~\cite{GeBeKr01}, the penDigits dataset from UCI database. The Extended Yale B dataset contains 2414 front-face images from 38 persons. In the experiment, we randomly select 8 persons, crop their images to $48\times 42$ and further reduce the dimension by projecting to the first $80$ principal components. As suggest in~\cite[page~12]{DBLP:journals/corr/abs-1010-3460}, we apply a crude whitening process to the projected data (removing the top two principal components) before applying FLS and LBF. We report SSC and SSSC without such whitening since they worked better when all $80$ directions are included. The penDigits dataset contains 7494 data points of 16 features, each of which represents a digit from 0 to 9. The results are reported in Table~\ref{table:3sets}. 

\begin{table}[htb!]\centering
  \caption{Average accuracy (time (s)).}\label{table:3sets} 
  \hfill \\
  \ra{1.3}
  \begin{tabular}{@{}ccccc@{}}\toprule
     & FLS & LBF & SSC & SSSC \\
    \midrule 
    ExtendYB & 0.68 ({\bf 1.79}) & 0.63 (4.36) & {\bf 0.87} (259.66) & 0.57 (15.02) \\
    penDigits & {\bf 0.68} (18.58) & 0.53 ({\bf 13.33}) & N/A & 0.51 (44.07) \\
    \bottomrule
  \end{tabular}
\end{table}

\section{Conclusion}
In this paper we introduce an efficient framework to approximate a large class of kernels and consequently obtain faster clustering algorithms. In the context of clustering, this framework proposes a novel procedure for fast subspace clustering. Its complexity is $O(KnD)$ for $n$ points with $K$ clusters and $D$ landmarks. Our theoretical analysis establishes such complexity for clustering algorithms associated with our framework. More importantly it bounds the $L_2$-error between the original and approximated eigenvectors of kernel matrices, which applies to FLS, Fourier random features and other landmark-based methods.

{\small
\bibliography{biblio}{}
\bibliographystyle{plain}
}

\appendix
%\subsection{Proof of Lemma~\ref{lm:rotation}}
\section{Proof of Lemma~3.1}

Let $(x_1, x_2)$ and $(x'_1, x'_2)$ be two pairs on $\Sb^d$ such that the angles $\theta(x_1,x_2) = \theta(x'_1, x'_2)$. Then there exists an orthogonal transformation $\mb{R}$ such that $x'_1 = \mb{R}x_1$ and $x'_2 = \mb{R}x_2$. Thus,
\begin{equation*}
\begin{aligned}
k(x'_1, x'_2) &= \int_{L \in Y} f(\mb{R}x_1, L) f(\mb{R}x_2, L) d\mu_Y(L) \\
 &= \int_{L \in Y} f(x_1, \mb{R}^{-1}L) f(x_2, \mb{R}^{-1}L) d\mu_Y(L) \\
 &= \int_{L \in Y} f(x_1, L) f(x_2, L) d\mu_Y(\mb{R}L) \\
 &= \int_{L \in Y} f(x_1, L) f(x_2, L) d\mu_Y(L) = k(x_1, x_2). 
\end{aligned}
\end{equation*}
We use the fact that $\mu_Y$ is uniform in the fourth equality. This equality means that $k(x_1, x_2) = g(\theta(x_1, x_2))$. We conclude the proof by noting that $\theta(x_1, x_2)$ and $\dist(x_1, x_2)$ uniquely determine each other for any pair $(x_1, x_2)$ on $\Sb^d$.

%\subsection{Proof of Theorem~\ref{thm:D}}
\section{Proof of Theorem~4.1}

We denote $h(x_1, x_2) = \psi(x_1)\psi(x_2) - k(x_1, x_2)$. Let $L_f = \|\nabla h(x_1^*, x_2^*)\|$, where $(x_1^*, x_2^*)=\mathrm{arg\, max}_{X\times X} \|\nabla h(x_1, x_2)\|$. Then
$$
\E L_f^2 \leq \E\|\nabla(\psi(x_1)\psi(x_2))\|^2 \leq C_f^4.
$$
The first inequality follows from the same argument in~\cite{DBLP:conf/nips/RahimiR07}. By Markov's inequality,
\begin{equation}\label{eq:markov}
\Pb( L_f \geq \frac{\epsilon}{2r} ) \leq \left( \frac{2r C_f^2}{\epsilon} \right)^2.
\end{equation}

On the other hand, if $\MM$ is a $d$-dimensional manifold with diameter $\mathrm{diam}(\MM)$, then $\MM\times \MM$ has dimension $2d$, diameter $\sqrt{2}\mathrm{diam}(\MM)$ and $r$-net covering number $T = \left( \frac{12\sqrt{2}d\mathrm{diam}(\MM)}{r} \right)^{4d}$. If $(x_1^{(i)}, x_2^{(i)})$ be the vertices of the $r$-net, then by Hoeffding's inequality
\begin{equation}\label{eq:hoeffding}
\Pb[\cup_{i=1}^T\{|h(x_1^{(i)}, x_2^{(i)})| \geq \epsilon/2\}] \leq 2T\exp(-D\epsilon^2/8).
\end{equation}
Equations~\eqref{eq:markov},~\eqref{eq:hoeffding} imply that
$$
\Pb[\sup_{\MM\times \MM} |h(x_1, x_2)| \leq \epsilon] \geq 1 - 2T\exp(-D\epsilon^2/8) - \left( \frac{2r C_f^2}{\epsilon} \right)^2.
$$
If we pick $r$ such that the second and third terms on the right-hand side are equal and simplifying the expression, then
$$
\Pb[\sup_{\MM\times \MM} |h(x_1, x_2)| \leq \epsilon] \geq 1 - C_{d, f}\exp(-D\epsilon^2/(16d+8))/\epsilon^2,
$$
where $C_{d, f}$ is a constant depending on $d$ and the bound of $f$ and its derivatives.
%where $C(d, f)$ is the constant $4608d^2\mathrm{\MM}^2 C_f^4$.

%\subsection{Proof of Lemma~\ref{lm:l2}}
\section{Proof of Lemma~4.3}

We first bound the $L_2$-norms of each factor in $H_i$.

\begin{equation}\label{eq:l2proof}
\begin{aligned}
\| \mb{D}^{-1/2} \|_2 &= \max_{i} D_{ii}^{-1/2} = \left( \min_{i} D_{ii} \right)^{-1/2} \leq (nl)^{-1/2}, \\
\| \hat{\mb{D}}^{-1/2} \|_2 & = \left( \min_{i} \hat{D}_{ii} \right)^{-1/2} \leq (n(1-\delta)l)^{-1/2}, \\
\|\mb{W}\|_2 & \leq n \max_{i,j} |W_{ij}| \leq nu, \quad \|\hat{\mb{W}}\|_2 \leq nu+n\delta l\\
\|\mb{W} - \hat{\mb{W}}\|_2 & \leq n \max_{i,j} |W_{ij} - \hat{W}_{ij}| \leq n \delta l, \\
\|\mb{D}^{-1/2} - \hat{\mb{D}}^{-1/2}\|_2 &= \max_{i} |D_{ii}^{-1/2} - \hat{D}_{ii}^{-1/2}| \leq \frac{ \max_{i} |D_{ii} - \hat{D}_{ii}|}{2(n(1-\delta)l)^{3/2}} \\
&\leq \frac{1}{2} \delta (1-\delta)^{-3/2} (nl)^{-1/2}.
\end{aligned}
\end{equation}
Here is a hint for the last inequality of~\eqref{eq:l2proof}. The function $f(x) = x^{-1/2}$ is a convex decreasing function over $x>0$. Therefore, $|f(x) - f(y)| \leq f'(\min\{x,y\}) |x-y| $. Then the last inequality in~\eqref{eq:l2proof} follows from the fact that
$$
\min_i \{\min\{D_{ii}, \hat{D}_{ii}\}\} \geq n(1-\delta)l.
$$
The upper bounds of $\|H_i\|_2$ follow immediately from~\eqref{eq:l2proof} and the inequality $\|AB\|_2 \leq \|A\|_2 \|B\|_2$ for any matrices $A, B$.

%\subsection{Proof of Lemma~\ref{lm:firsteig}}
\section{Proof of Lemma~4.4}

Let the linear operator $T$ on $C(B(\mb{0},1))$ be defined as follows:
$$
T(f) (\mb{x}) =  \int_{\mb{y}\in B(\mb{0}, 1)} k(\mb{x}, \mb{y}) f(\mb{y})/\sqrt{d(\mb{x}) d(\mb{y})} dP(\mb{y}),
$$
where the degree function $d(\mb{x}) = \int_{\mb{y}\in B(\mb{0},1)}  k(\mb{x}, \mb{y})  dP(\mb{y})$. Since $T$ is a compact operator, its eigenvalue accumulation point is $0$. It is easy to see that $1$ is an eigenvalue of $T$ with the eigenvector $\sqrt{d(\mb{y})}$. Luxburg et al.~\cite{MR2396807} showed that the eigenvalues and eigenvectors of $\mb{L}$ converge to those of $T$ with a convergence rate $O(\frac{1}{\sqrt{n}})$. Since $\mb{L}$ has all its eigenvalues in $[0,1]$, so is $T$. We summerize this in Lemma~\ref{lm:U}

\begin{lemma}\label{lm:U}
$T$ has all eigenvalues between $[0,1]$. $1$ is its eigenvalue. The only accumulation point of the eigenvalues is $0$.
\end{lemma}

Let $\sigma(T)$ and $\lambda_1>0$ be the spectrum and the second largest eigenvalue of $T$ respectively. For simplicity, we assume $\lambda_1$ ($\lambda_{1,n}$) is a simple eigenvalue. Lemma~\ref{lm:U} implies the distance 
$$
\dist(\lambda_1, \{0\}\cup \sigma(U) \backslash \{\lambda_1\}) = 2c > 0
$$ 
for some $\delta>0$. Theorem~15 in~\cite{MR2396807} implies that for any constant probability $p$, there exists $N_p$ such that if $n > N_p$, the 
$$
\dist(\lambda_{1,n}, \{0\}\cup \sigma(\mb{L}_n) \backslash \{\lambda_{1,n}\}) \geq c
$$
with probability at least $p$.

\end{document}